\documentclass[10pt,twocolumn,letterpaper]{article}

\usepackage{wacv}
\usepackage{times}
\usepackage{epsfig}
\usepackage{graphicx}
\usepackage{amsmath}
\usepackage{amssymb}
\usepackage{subcaption}
\usepackage{caption}
\usepackage{multirow}
\graphicspath{{./figures/}}
\usepackage{arydshln}
\usepackage{color}
\usepackage{cite}
\usepackage{animate}
\usepackage{gensymb}

\def\wacvPaperID{0011} 

\wacvfinalcopy 

\ifwacvfinal
\def\assignedStartPage{001} 
\fi

\ifwacvfinal
\usepackage[breaklinks=true,bookmarks=false]{hyperref}
\else
\usepackage[pagebackref=true,breaklinks=true,colorlinks,bookmarks=false]{hyperref}
\fi

\ifwacvfinal
\else
\fi

\begin{document}

\title{Multi-View Motion Synthesis via Applying Rotated Dual-Pixel Blur Kernels} 

\author{Abdullah Abuolaim \qquad Mahmoud Afifi \qquad Michael S. Brown\\
York University\\
{\tt\small \{abuolaim,mafifi,mbrown\}@eecs.yorku.ca}
}

\maketitle

\begin{abstract}
Portrait mode is widely available on smartphone cameras to provide an enhanced photographic experience. One of the primary effects applied to images captured in portrait mode is a synthetic shallow depth of field (DoF). The synthetic  DoF (or bokeh effect) selectively blurs regions in the image to emulate the effect of using a large lens with a wide aperture. In addition, many applications now incorporate a new image motion attribute (NIMAT) to emulate background motion, where the motion is correlated with estimated depth at each pixel. In this work, we follow the trend of rendering the NIMAT effect by introducing a modification on the blur synthesis procedure in portrait mode. In particular, our modification enables a high-quality synthesis of multi-view bokeh from a single image by applying rotated blurring kernels. Given the synthesized multiple views, we can generate aesthetically realistic image motion similar to the NIMAT effect. We validate our approach qualitatively compared to the original NIMAT effect and other similar image motions, like Facebook 3D image. Our image motion demonstrates a smooth image view transition with fewer artifacts around the object boundary.
\end{abstract}

\begin{figure}[t]
\centering
\animategraphics[width=0.47\textwidth]{8}{figures/teaser_rotated/t_}{1}{24}
\vspace{-2mm}
\caption{This figure shows a comparison between different image motion effects. We also show the output of the traditional bokeh synthesis. Our approach takes the sharp image (i.e., deep DoF) to generate the image motion. Other approaches start with the blurry input (i.e., shallow DoF) to synthesize the image motion. \textbf{Note: this figure is animated; click on the image to start the animation. It is recommended to open this PDF in Adobe Acrobat Reader to work properly.}}\label{fig:teaser-rotated-kernels}
\vspace{-5mm}
\end{figure}

\section{Introduction}
Unlike digital single-lens reflex (DSLR) and mirrorless cameras, smartphone cameras cannot produce a natural shallow depth of field (DoF) due to the camera's small aperture and simple optical system.  Instead, many smartphones (e.g., iPhone 12, Google Pixel 4, Samsung Galaxy) emulate a shallow DoF via a {\it portrait mode} setting that processes the image at capture time.  These methods typically isolate the subject from the background and then blur the background to emulate the swallow DoF~\cite{wadhwa2018synthetic}.  An example is shown in the first row of Fig.~\ref{fig:teaser-rotated-kernels}.

\begin{figure*}[t]
\centering
\includegraphics[width=\linewidth]{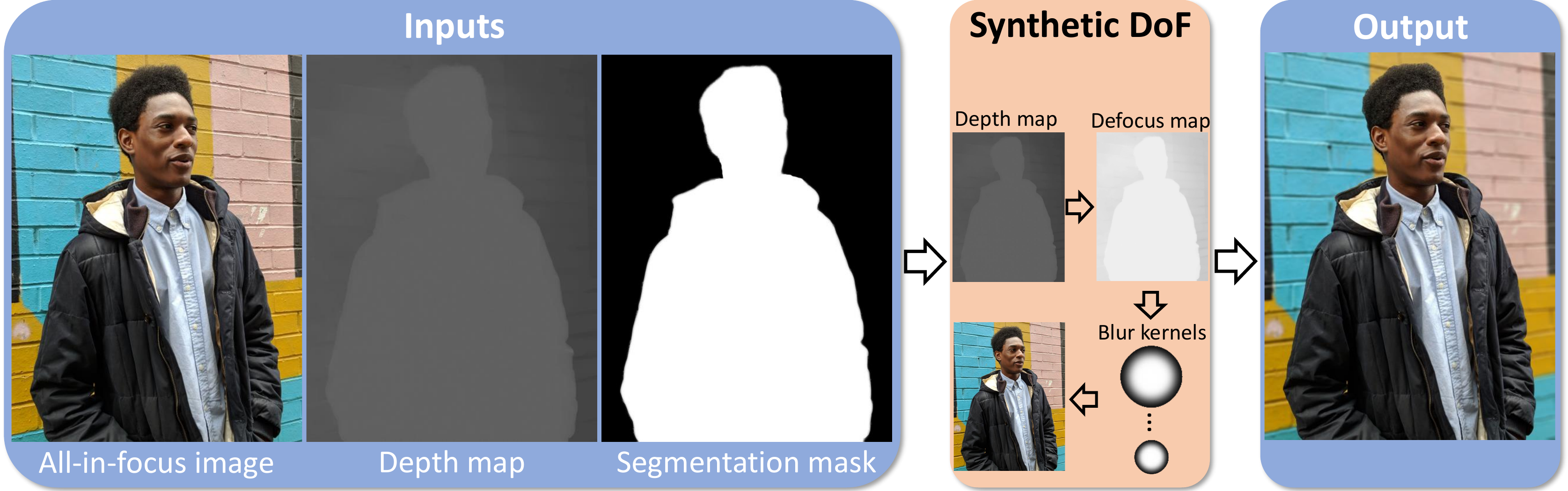}
\caption{This figure shows a typical synthetic shallow depth of field (DoF) processing framework. This framework takes three inputs: single image, estimated depth map, and segmentation mask. Given the inputs, the synthetic DoF unit produces the desired image. The image, depth map, and segmentation mask are taken from the dataset in~\cite{wadhwa2018synthetic}. }\label{fig:synth_dof_framework}
\end{figure*}

Most smartphone cameras apply the synthetic bokeh effect using a common image processing framework. This traditional procedure takes an input image with minimal DoF blur and an estimated depth map to determine the blur kernel size at each pixel (i.e., defocus map). In some cases, a segmentation mask is also used to avoid blurring pixels that belong to the people and their accessories. Fig.~\ref{fig:synth_dof_framework} shows an illustrative example of the common synthetic bokeh framework.

Recently, Abuolaim et al. proposed a new image motion attribute (NIMAT) effect~\cite{abuolaim2021multi} that generates multiple sub-aperture views based on DoF blur and dual-pixel (DP) image formation. Abuolaim et al.'s method produces multiple views from a single input image captured by a DSLR camera and has a natural shallow DoF. Their DP- and DoF-based view synthesis is designed to generate pixel motion correlated to the defocus blur size at each pixel. However, obtaining an image with a natural shallow DoF using a smartphone camera is difficult, as mentioned earlier. Inspired by NIMAT~\cite{abuolaim2021multi}, we provide a similar effect by modifying the traditional synthetic bokeh framework. Our modification enables synthesizing shallow DoF along with generating multiple views by applying a rotated blurring kernel. In our proposed framework, the defocus blur kernel shape is determined based on the sub-aperture image formation found in DP sensors. To our knowledge, we are the first to introduce this novel synthetic bokeh and DP-/DoF-based multi-view synthesis. Fig.~\ref{fig:teaser-rotated-kernels} shows a comparison of different image motion approaches. It also provides the output of the traditional bokeh synthesis in the first row. Recall that other image motion approaches do not synthesize the bokeh effect.  As a result, our method combines image motion and synthetic DoF into a single step.  As demonstrated in Fig.~\ref{fig:teaser-rotated-kernels}, our image motion exhibits a smooth view transition with fewer artifacts around the object boundary compared to other approaches.

\section{Related Work}
\paragraph{Synthetic bokeh}
The bokeh effect in photography is an aesthetic quality of the blur that renders the main subject of the taken photo in focus while the background details fall out of focus. As mentioned earlier, standard smartphone cameras cannot produce such bokeh photographs due to the small size of the aperture and short focal length used in almost all smartphone cameras. Due to this limitation, a large body of work has targeted ways to emulate a shallow DoF image for smartphone cameras (e.g., \cite{hernandez2014lens, ha2016high, tang2017depth, yu20143d, suwajanakorn2015depth, ignatov2020rendering,wadhwa2018synthetic}).

Prior methods require either up-down translation of the camera (e.g., \cite{hernandez2014lens}) or benefits from the parallax caused by accidental handshake during capturing (e.g., \cite{ha2016high, yu20143d}). However, both strategies may lead to undesirable results as they rely on a specific type of movement that is not always applied in real scenarios. As a result, having low parallax limits these methods' ability to work properly.

Another strategy requires multi-image capturing, or stereo imaging, to estimate image depth from defocus cues extracted from these multiple images, or stereo pairs, of the same scene \cite{suwajanakorn2015depth, tang2017depth, garg2016unsupervised, godard2017unsupervised, zhou2017unsupervised, xie2016deep3d}. However, this strategy results in ghosting effects and cannot work properly with non-static objects.

Instead of relying on multi-image capture, monocular single-image depth estimation methods are adopted to predict depth information using either inverse rendering \cite{barron2014shape, horn1975obtaining} or supervised machine learning ~\cite{eigen2014depth, hoiem2005automatic, liu2015learning, saxena2007learning}. Given the estimated depth map, synthetic rendering of shallow DoF images is then a straightforward process. However, the quality of this synthetic bokeh effect is tied to the accuracy of the estimated depth map. In recent years, learning-based depth estimation methods have achieved impressive results; however, like most deep learning-based techniques, such learning depth estimators often suffer from poor generalization to images taken under conditions beyond training examples. Thus, synthesized shallow DoF images could suffer from obvious artifacts around the main object's edges.

\begin{figure*}[t]
\centering
\includegraphics[width=\linewidth]{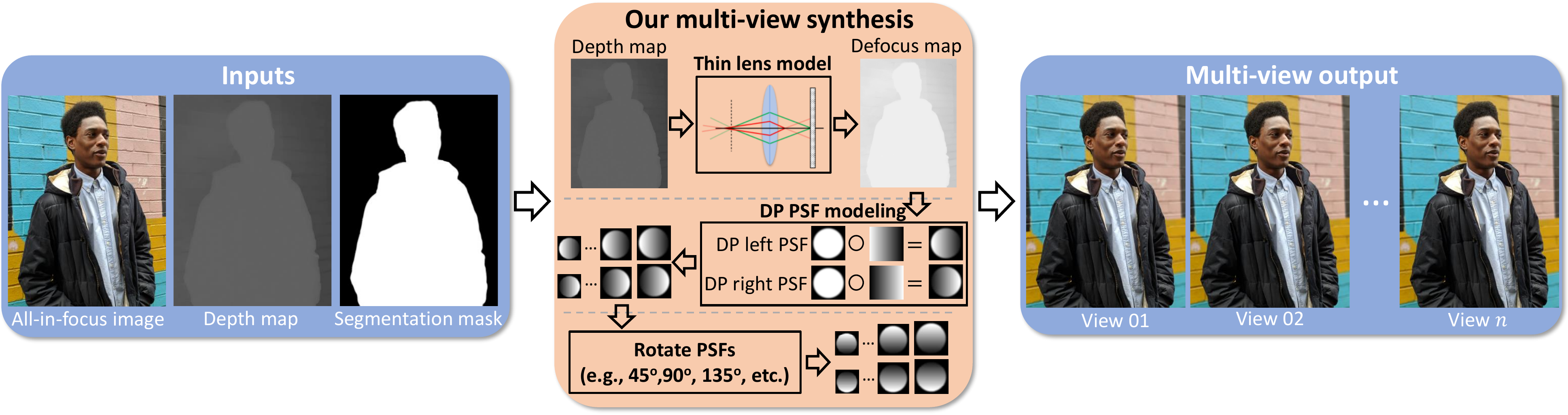}
\caption{An overview of our proposed framework for multi-view synthesis based on rotated DP blur kernels. This framework takes three inputs: single image, estimated depth map, and segmentation mask. Given the inputs, the multi-view synthesis unit produces $n$ views based on the number of rotated point spread functions (PSFs). The image, depth map, and segmentation mask are taken from the dataset in~\cite{wadhwa2018synthetic}. }\label{fig:synth_dof_framework_ours}
\end{figure*}

To mitigate failure cases in single-image depth estimation, a few methods propose to replace the depth estimation process with some constraints in the scene to improve the results. For example, by dealing only with photos of people against a distant background, bokeh effects can be generated without a need for a depth map estimation \cite{shen2016automatic, shen2016deep}. With this reasonable constraint, synthetic shallow DoF can be achieved by first segmenting out the human subject. This is typically performed using a trained convolutional neural network.  Next, the background can be blurred using a global blur kernel. While effective, this approach assumes a constant difference in depth between the main subject (i.e., people) and the background. In addition, this approach requires a deep network to segment people from images properly.

Unlike all methods above, in this paper, our goal is to produce an image motion effect similar to the NIMAT effect~\cite{abuolaim2021multi}. A high-quality bokeh synthesis is an extra by-product output.

\paragraph{DP sensor}
DP sensors were developed as a means to improve the camera's autofocus system. The DP design produces two sub-aperture views of the scene that exhibit differences in phase that are correlated to the amount of defocus blur. Then, the phase difference between the left and right sub-aperture views of the primary lens is calculated to measure the blur amount. The phase information is also used to adjust the camera's lens such that the blur is minimized. While intended for autofocus~\cite{abuolaim2018revisiting,abuolaim2020online}, the DP images have been found useful for other tasks, such as depth map estimation~\cite{punnappurath2020modeling,garg2019learning,zhang20202}, defocus deblurring~\cite{abuolaim2020defocus,abuolaim2021learning,abuolaim2021ntire}, and synthetic DoF~\cite{wadhwa2018synthetic}.

\section{Defocus-Based Multi-View Synthesis}

In this section, we describe our framework for multi-view synthesis based on rotated DP blur kernels. An overview of the proposed framework is shown in Fig.~\ref{fig:synth_dof_framework_ours}. First, we introduce the thin lens model used to determine the blur kernel size at each pixel. Then, the DP point spread function (PSF) is described in Sec.~\ref{sec:dpPSF}. Afterward, Sec.~\ref{sec:applyBlur} introduces the defocus blur procedure. Lastly, Sec.~\ref{sec:multiView} explains the process of multi-view synthesis via rotated PSFs.


\subsection{PSF Size Based on the Thin Lens Model\label{sec:thinLens}}
\begin{figure}[t]
\centering
\includegraphics[width=\linewidth]{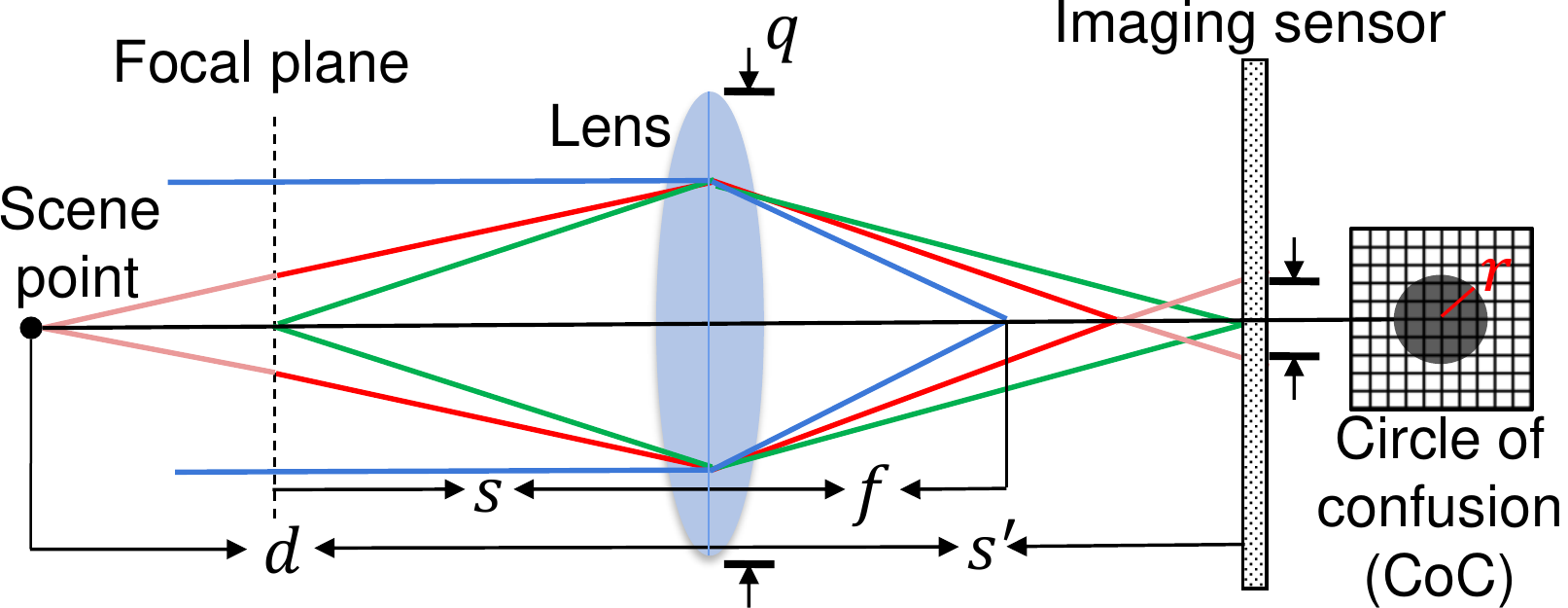}
\caption{Thin lens model illustration and dual-pixel (DP) image formation. The circle of confusion (CoC) size is calculated for a given scene point using its distance from the lens, camera focal length, and aperture size. Note: we acknowledge that this figure was adapted from~\cite{abuolaim2021learning}}.\label{fig:thinlens}
\end{figure}

\begin{figure*}[t]
    \centering
    \includegraphics[width=\linewidth]{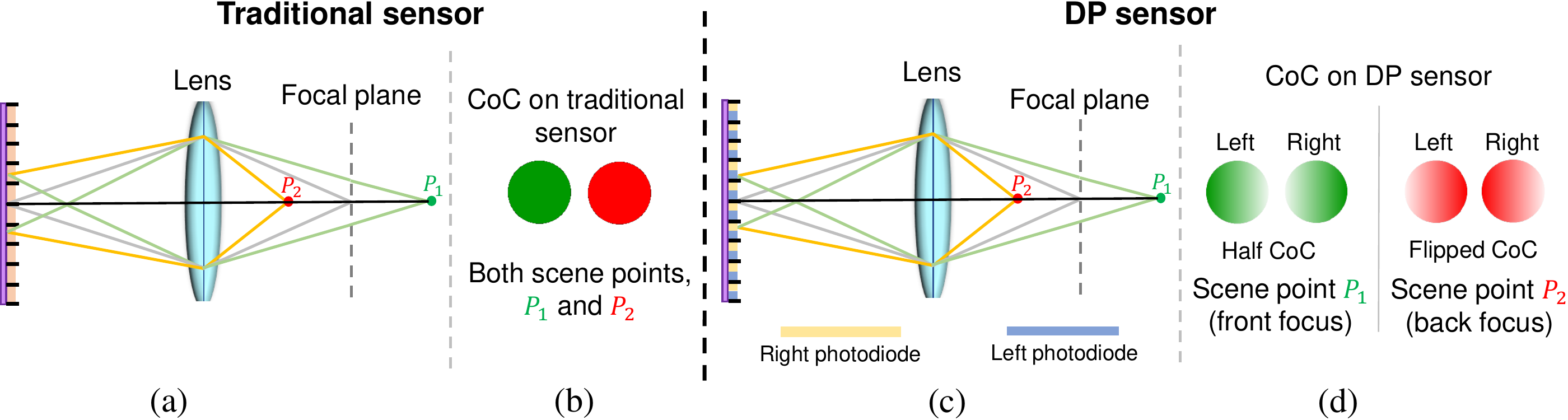}
    \caption{Circle of confusion (CoC) formation in DP sensors. (a) Traditional sensor and (c) DP sensor. (b) and (d) are the CoC formation on the 2D imaging sensor of two scene points, $P1$ and $P2$. On the two DP views, the half-CoC flips direction if the scene point is in front or back of the focal plane. Note: we acknowledge that this figure was adapted from~\cite{abuolaim2021multi}.}
    \label{fig:dp_formation}
\end{figure*}

The size of the PSFs at each pixel in the image can be calculated using the depth map. Therefore, we model camera optics using a thin lens model that assumes negligible lens thickness, helping to simplify optical ray tracing calculations~\cite{potmesil1981lens}. This model can approximate the circle of confusion (CoC) size for a given point based on its distance from the lens and camera parameters (i.e., focal length, aperture size, and focus distance). This model is illustrated in Fig.~\ref{fig:thinlens}, where $f$ is the focal length, $s$ is the focus distance, and $d$ is the distance between the scene point and camera lens. The distance between the lens and sensor $s'$, and the aperture diameter $q$ are defined as:
\begin{equation}\label{eq:lensDistance}
s'=\frac{f\;s}{s-f},
\end{equation}
\begin{equation}\label{eq:lensAperture}
q=\frac{f}{F},
\end{equation}

\noindent where $F$ is the f-number ratio. Then, the CoC radius $r$ of a scene point located at distance $d$ from the camera is:
\begin{equation}\label{eq:cocRadius}
r=\frac{q}{2} \times \frac{s'}{s} \times \frac{d-s}{d}.
\end{equation}


\subsection{PSF Shape Based on DP Image Formation\label{sec:dpPSF}}

Once the radius of the PSF is calculated at each pixel (Sec.~\ref{sec:thinLens}), we need to decide the PSF shape to be applied. In this section, we adopt a DP-based PSF shape for DP view synthesis.

We start with a brief overview of DP sensors.  A DP sensor uses two photodiodes at each pixel location with a microlens placed on the top of each pixel site, as shown in Fig.~\ref{fig:dp_formation}-c. This design was developed by Canon to improve camera autofocus by functioning as a simple two-sample light field camera. The two-sample light-field provides two sub-aperture views of the scene and, depending on the sensor's orientation, the views can be referred to as left/right or top/down pairs; we follow the convention of prior papers~\cite{abuolaim2020defocus,punnappurath2020modeling} and refer to them as the left/right pair. The light rays coming from scene points that are within the camera's DoF exhibit little to no difference in phase between the views. On the other hand, light rays coming from scene points outside the camera's DoF exhibit a noticeable defocus disparity in the {\it left}-{\it right} views. The amount of defocus disparity is correlated to the amount of defocus blur.

\begin{figure}[t]
     \centering
     \begin{subfigure}[b]{0.23\textwidth}
         \centering
         \includegraphics[width=\textwidth]{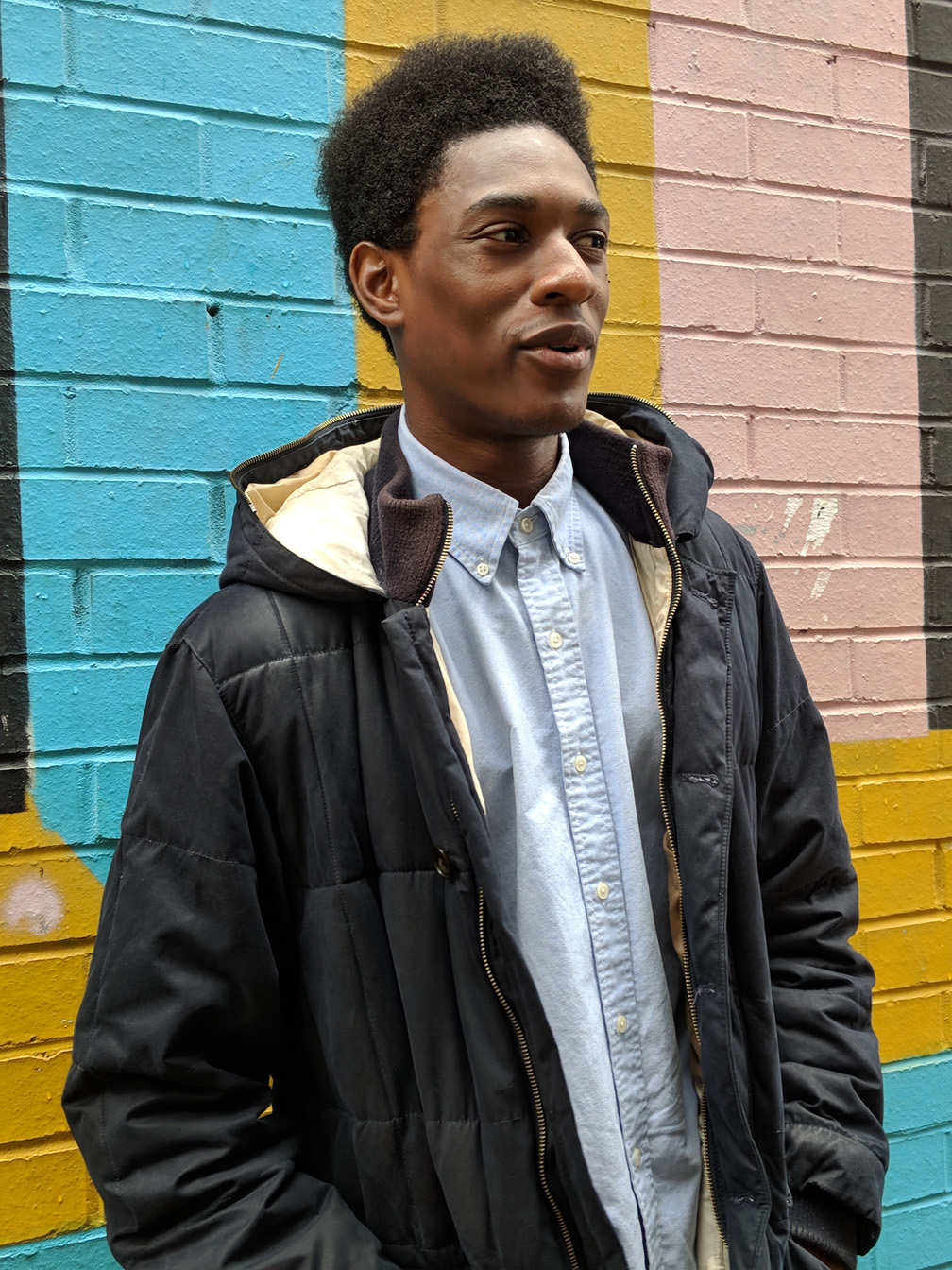}
         \caption{All-in-focus input}
     \end{subfigure}
     \begin{subfigure}[b]{0.23\textwidth}
         \centering
         \includegraphics[width=\textwidth]{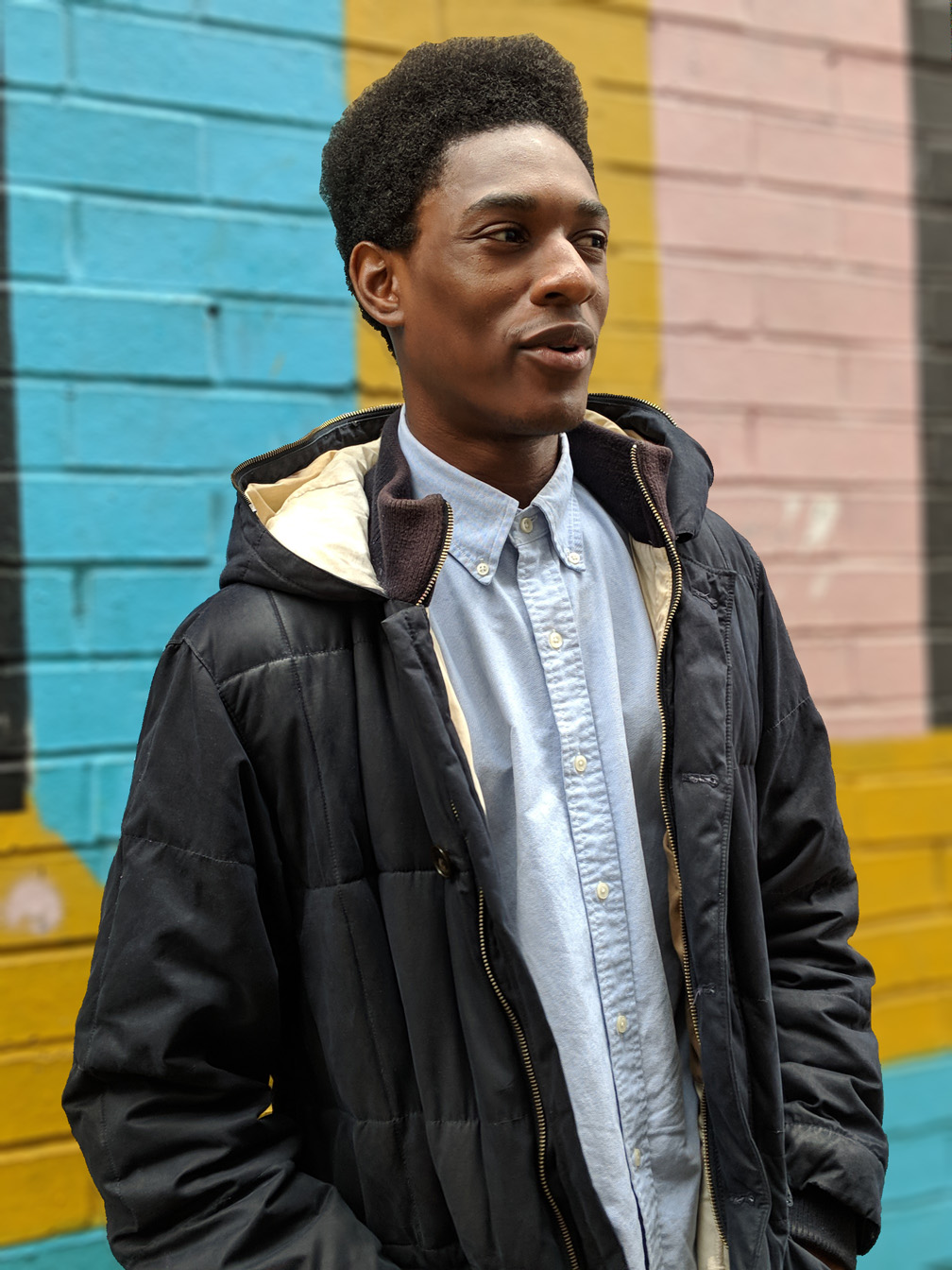}
         \caption{Our synthetic bokeh}
     \end{subfigure}

    \vspace{2mm}
    
     \begin{subfigure}[b]{0.23\textwidth}
         \centering
         \includegraphics[width=\textwidth]{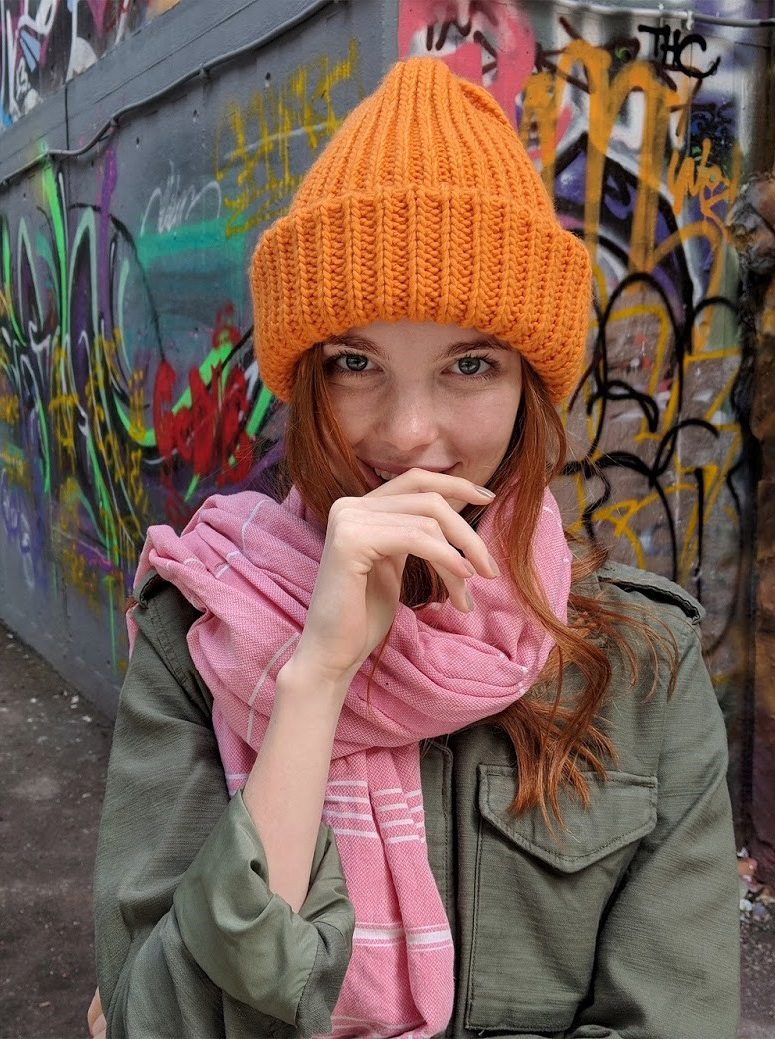}
         \caption{All-in-focus input}
     \end{subfigure}
     \begin{subfigure}[b]{0.23\textwidth}
         \centering
         \includegraphics[width=\textwidth]{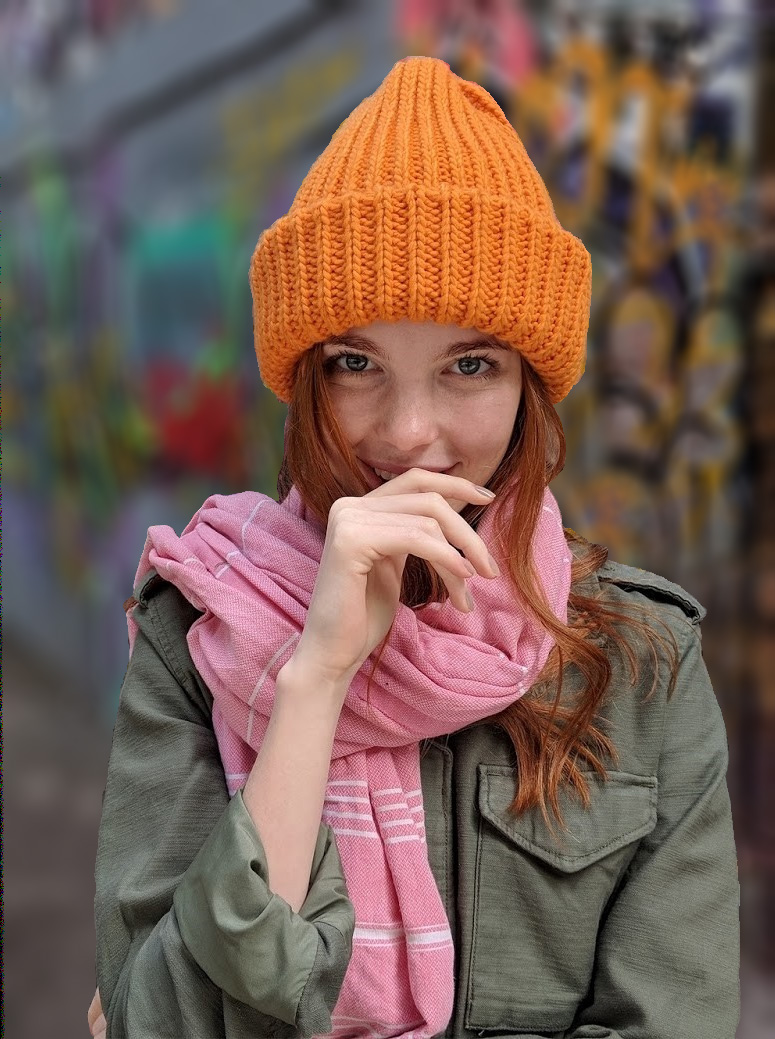}
         \caption{Our synthetic bokeh}
     \end{subfigure}
    \caption{Our synthetic bokeh results given an input all-in-focus image. The images used in this figure are from the synthetic DoF dataset~\cite{wadhwa2018synthetic}.}
    \label{fig:syntheticBokeh}
\end{figure}

\begin{figure*}[t]
     \centering
     \begin{subfigure}[b]{0.49\textwidth}
         \centering
         \animategraphics[trim=0 400 0 122, width=\textwidth]{3}{figures/dp_ours/ours_}{1}{12}
         \caption{Our synthetic DP views}
         \label{fig:teaser_our_dp}
     \end{subfigure}
    \begin{subfigure}[b]{0.49\textwidth}
          \centering
          \animategraphics[width=\textwidth]{3}{figures/dp_gt/5_g_}{1}{12}
          \caption{Real DP views}
          \label{fig:teaser_gt_dp}
	\end{subfigure}
    \caption{Results from our DP-view synthesis framework based on defocus blur in DP sensors. (a) Our synthetic DP views. (b) Real DP views. Our framework can produce DP views that have defocus disparity similar to the one found in real DP sensors. The image on the left is from the synthetic DoF dataset~\cite{wadhwa2018synthetic}. \textbf{Note: the DP views are animated; click on the image to start the animation. It is recommended to open this PDF in Adobe Acrobat Reader to work properly.}
    }
    \label{fig:dpDtVsSynth}
\end{figure*}

Unlike traditional stereo, the difference between the DP views can be modeled as the latent sharp image being blurred in two different directions using a half-circle PSF~\cite{punnappurath2020modeling}. This is illustrated in the resultant CoC of Fig.~\ref{fig:dp_formation}-d. The ideal case of a half-circle CoC on real DP sensors is only an approximation due to constraints of the sensor's construction and lens array. These constraints allow a part of the light ray bundle to leak into the other-half dual pixels (see half CoC of left/right views in Fig.~\ref{fig:dp_formation}-d).

Unlike other approaches~\cite{abuolaim2021learning,punnappurath2020modeling}, we provide a simplified model of the DP PSF using a disk $\mathbf{C}$ shape that is element-wise multiplied by a ramp mask as follows:
\begin{equation} \label{equ:dpPsf}
\mathbf{H}_l = \mathbf{C} \circ \mathbf{M}_l, \quad \mbox{s.t. } \ \mathbf{H}_l \geq \mathbf{0},\; \mbox{with} \sum \mathbf{H}_l = 1,
\end{equation}
where $\circ$ denotes element-wise multiplication, $\mathbf{M}_l$ is a 2D ramp mask with a constant intensity fall-off towards the right direction, and $\mathbf{H}_l$ is the left DP PSF. One interesting property of the DP sensors is that the right DP PSF $\mathbf{H}_r$ is the $\mathbf{H}_l$ that is flipped around the vertical axis -- namely, $\mathbf{H}_l^f$:
\begin{equation} \label{equ:dpLR}
\mathbf{H}_r = \mathbf{H}_l^f.
\end{equation}

Another interesting property of the DP PSFs is that the orientation of the ``half CoC'' of each left/right view reveals if the scene point is in front or back of the focal plane~\cite{punnappurath2020modeling,abuolaim2021learning,abuolaim2021multi}. Following the prior work of modeling directional blur using DP image formation, we also select the DP-based “half CoC” PSF model to capture the directional blur in this paper. However, this directional blur PSF does not have to be DP-based and can be any generic PSF that involves blurring and shifting the image content. Therefore, we test other non-DP-based directional PSF in Sec.~\ref{subsec:other-psfs}.

\subsection{Applying Synthetic Defocus Blur\label{sec:applyBlur}}

In our framework, we use an estimated depth map to apply synthetic defocus blur in the process of generating a shallow DoF image. To blur an image based on the computed CoC radius $r$, we first decompose the image into discrete layers according to per-pixel depth values, where the maximum number of layers is set to $500$ (similar to~\cite{lee2019deep}). Then, we convolve each layer with the DP PSF (Sec.~\ref{equ:dpPsf}), blurring both the image and mask of the depth layer. Next, we compose the blurred layer images in order of back-to-front, using the blurred masks. For an all-in-focus input image $\mathbf{I}_s$, we generate two images -- namely, the \emph{left} $\mathbf{I}_l$ and \emph{right} $\mathbf{I}_r$ sub-aperture DP views -- as follows (for simplicity, let $\mathbf{I}_s$ be a patch with all pixels from the same depth layer):
\begin{equation} \label{eqn:dpLView}
\mathbf{I}_l = \mathbf{I}_s \ast \mathbf{H}_l,
\end{equation}
\begin{equation} \label{eqn:dpRView}
\mathbf{I}_r = \mathbf{I}_s \ast \mathbf{H}_r,
\end{equation}
where $\ast$ denotes the convolution operation. The final output image $\mathbf{I}_b$ (i.e., synthetic shallow DoF image) that is produced by the traditional {\it portrait mode} can be obtained as follows:
\begin{equation} \label{eqn:dpCombined}
\mathbf{I}_b = \frac{\mathbf{I}_l + \mathbf{I}_r}{2}.
\end{equation}

Fig.~\ref{fig:syntheticBokeh} shows the results of the generated synthetic bokeh image $\mathbf{I}_b$ using our proposed framework. Furthermore, our synthetically generated DP views exhibit defocus disparity similar to what we find in real DP data, where the in-focus regions show no disparity and the out-of-focus regions have defocus disparity. We provide in Fig.~\ref{fig:dpDtVsSynth} an animated comparison between our generated DP views and real DP views extracted from a Canon DSLR camera.

\begin{figure*}[t!]
     \centering
     \begin{subfigure}[b]{0.325\textwidth}
         \centering
         \animategraphics[width=\textwidth]{8}{figures/fb_view_res_1/fb_}{1}{24}
         \caption{Facebook 3D image (click)}
     \end{subfigure}
     \begin{subfigure}[b]{0.325\textwidth}
         \centering
         \animategraphics[width=\textwidth]{8}{figures/nimat_res_1/r_}{1}{24}
         \caption{NIMAT effect~\cite{abuolaim2021multi} (click)}
     \end{subfigure}
     \begin{subfigure}[b]{0.325\textwidth}
         \centering
         \animategraphics[width=\textwidth]{8}{figures/mult_view_res_1/r_}{1}{24}
         \caption{Our NIMAT effect (click)}
     \end{subfigure}
     
    \vspace{5mm}
    
     \begin{subfigure}[b]{0.325\textwidth}
         \centering
         \animategraphics[width=\textwidth]{8}{figures/fb_view_res_2/fb_}{1}{24}
         \caption{Facebook 3D image (click)}
     \end{subfigure}
     \begin{subfigure}[b]{0.325\textwidth}
         \centering
         \animategraphics[width=\textwidth]{8}{figures/nimat_res_2/r_}{1}{24}
         \caption{NIMAT effect~\cite{abuolaim2021multi} (click)}
     \end{subfigure}
    \begin{subfigure}[b]{0.325\textwidth}
          \centering
          \animategraphics[width=\textwidth]{8}{figures/mult_view_res_2/r_}{1}{24}
          \caption{Our NIMAT effect (click)}
	\end{subfigure}
    \caption{A comparison between different image motion approaches. This image motion is produced by animating the synthetic output views of each approach. Two cases of scene depth variation are provided: a small depth variation in the first row and a large one in the second row. Our proposed image motion produces a pleasant motion transition and fewer artifacts compared to others. The images used in this figure are from the synthetic DoF dataset~\cite{wadhwa2018synthetic}. \textbf{Note: the synthetic output views are animated; click on the image to start the animation. It is recommended to open this PDF in Adobe Acrobat Reader to work properly.}
    }
    \label{fig:multi_view}
\end{figure*}

\subsection{Multi-View Synthesis \label{sec:multiView}}
The main idea of this work is to generate multiple views from an all-in-focus image with its corresponding depth map. Therefore, we can generate an aesthetically realistic image motion by synthesizing a multi-view version of a given single image. As discussed in Sec.~\ref{sec:dpPSF}, the DP two sub-aperture views of the scene depending on the sensor's orientation and, in this work, our formation contain left/right DP pairs, and consequently, our framework synthesizes the horizontal DP disparity as shown in Fig.~\ref{fig:dpDtVsSynth}. We can synthesize additional views with different ``DP disparity'' by rotating the PSFs during the multi-view synthesis process as shown in Fig.~\ref{fig:synth_dof_framework_ours}. For example, eight views can be generated by performing a $45^o$ clockwise rotation step three times (i.e., $45^o$, $90^o$, $135^o$). Then, we generate our effect by alternating the output views to produce the image motion.

\section{Experiments}
\subsection{Results Using DP PSF}
Following the qualitative comparison procedure introduced in~\cite{abuolaim2021multi}, we provide the animated image motion (or NIMAT effect) of different approaches in Fig.~ \ref{fig:multi_view}. In particular, we compare ours with the results from~\cite{abuolaim2021multi} and the Facebook 3D image. As mentioned earlier and unlike other approaches, our proposed framework starts with the deep DoF image (i.e., almost all-in-focus) to produce the synthetic bokeh (or synthetic shallow DoF) image and the multiple DoF/DP-based views. Therefore, we provide the synthetic bokeh image as input to other approaches. This section also introduces the NIMAT-like effect from the common Facebook 3D image by uploading a single image and rendering the 3D version. Then, we save multiple frames at different view directions following the circular pixel motion transition found in the NIMAT effect~\cite{abuolaim2021multi}.

The results in this section show two cases of scene depth variations –- namely, a small depth variation (Fig.~ \ref{fig:multi_view}, first row) and a large one (Fig.~ \ref{fig:multi_view}, second row). While the Facebook 3D image motion is sufficient in the first row, it suffers from few artifacts around the foreground object boundary (e.g., the wall behind the person's head and arm). As for the NIMAT effect results from~\cite{abuolaim2021multi} in the first row, the image motion is barely noticeable in the background due to the small blur size that is a result of the small scene depth variation.

The second row of Fig.~ \ref{fig:multi_view} shows the large depth variation case, where the blur size varies from small to large. In this case, the Facebook 3D image exhibits noticeable and unpleasing artifacts (e.g., missing pixels). While the NIMAT effect from~\cite{abuolaim2021multi} produces pleasing image motion, we can still spot few artifacts that do not exist in ours. Note that we are aware the Facebook 3D image is not made for the same purpose, but we rendered it with the same motion transition settings of the NIMAT effect for comparison purposes.
\begin{figure}[t!]
     \centering
     \begin{subfigure}[b]{0.234\textwidth}
          \centering
          \animategraphics[width=\textwidth]{8}{figures/mult_view_res_2/r_}{1}{24}
          \caption{NIMAT -- DP PSF (click)}
	 \end{subfigure}
     \begin{subfigure}[b]{0.234\textwidth}
          \centering
          \animategraphics[width=\textwidth]{8}{figures/mult_view_res_2_ramp/r_}{1}{24}
          \caption{NIMAT -- Ramp PSF (click)}
	 \end{subfigure}
    
     \begin{subfigure}[b]{0.234\textwidth}
         \centering
         \includegraphics[width=\textwidth]{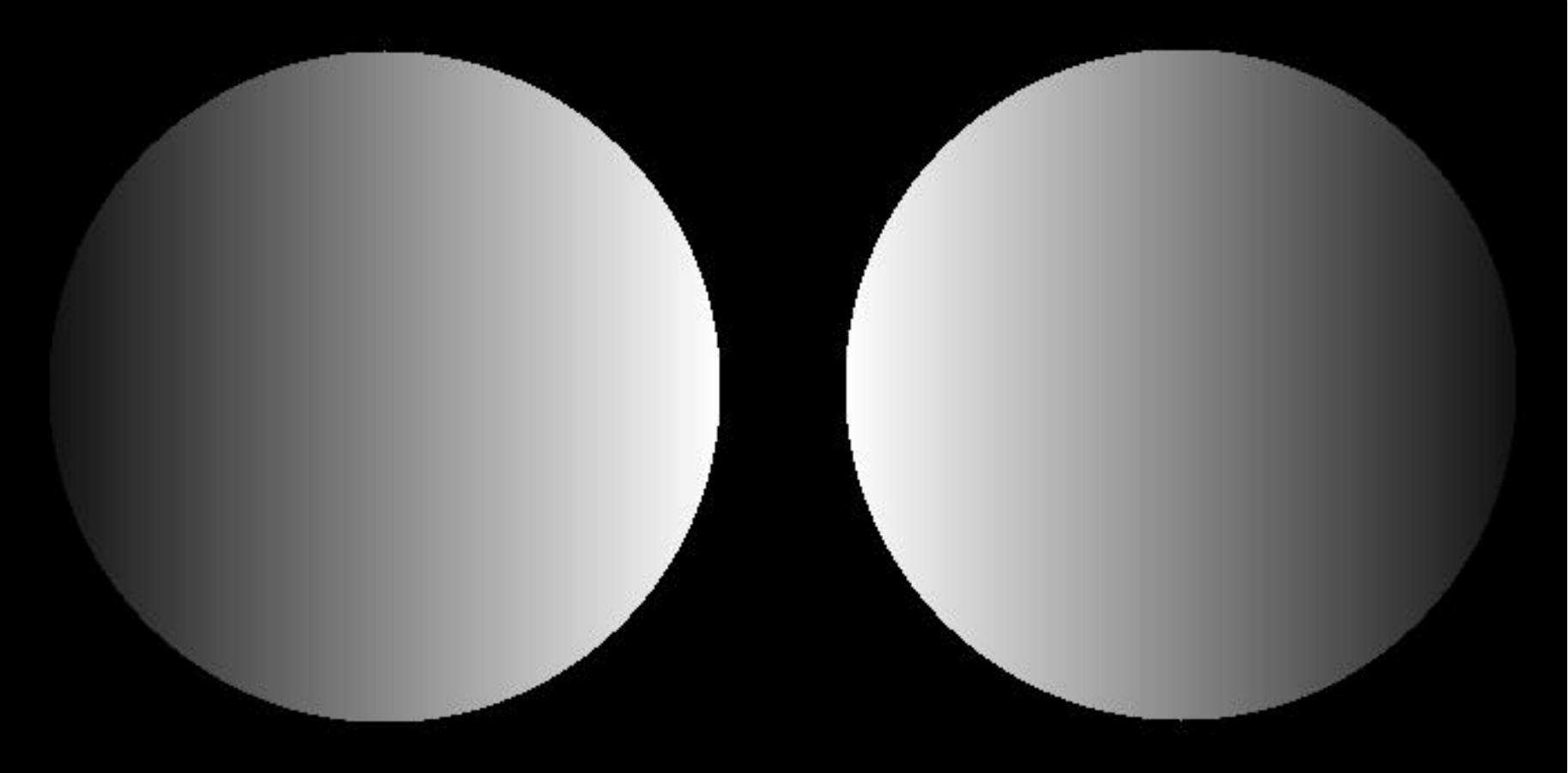}
         \caption{DP PSF}
     \end{subfigure}
     \begin{subfigure}[b]{0.234\textwidth}
         \centering
         \includegraphics[width=\textwidth]{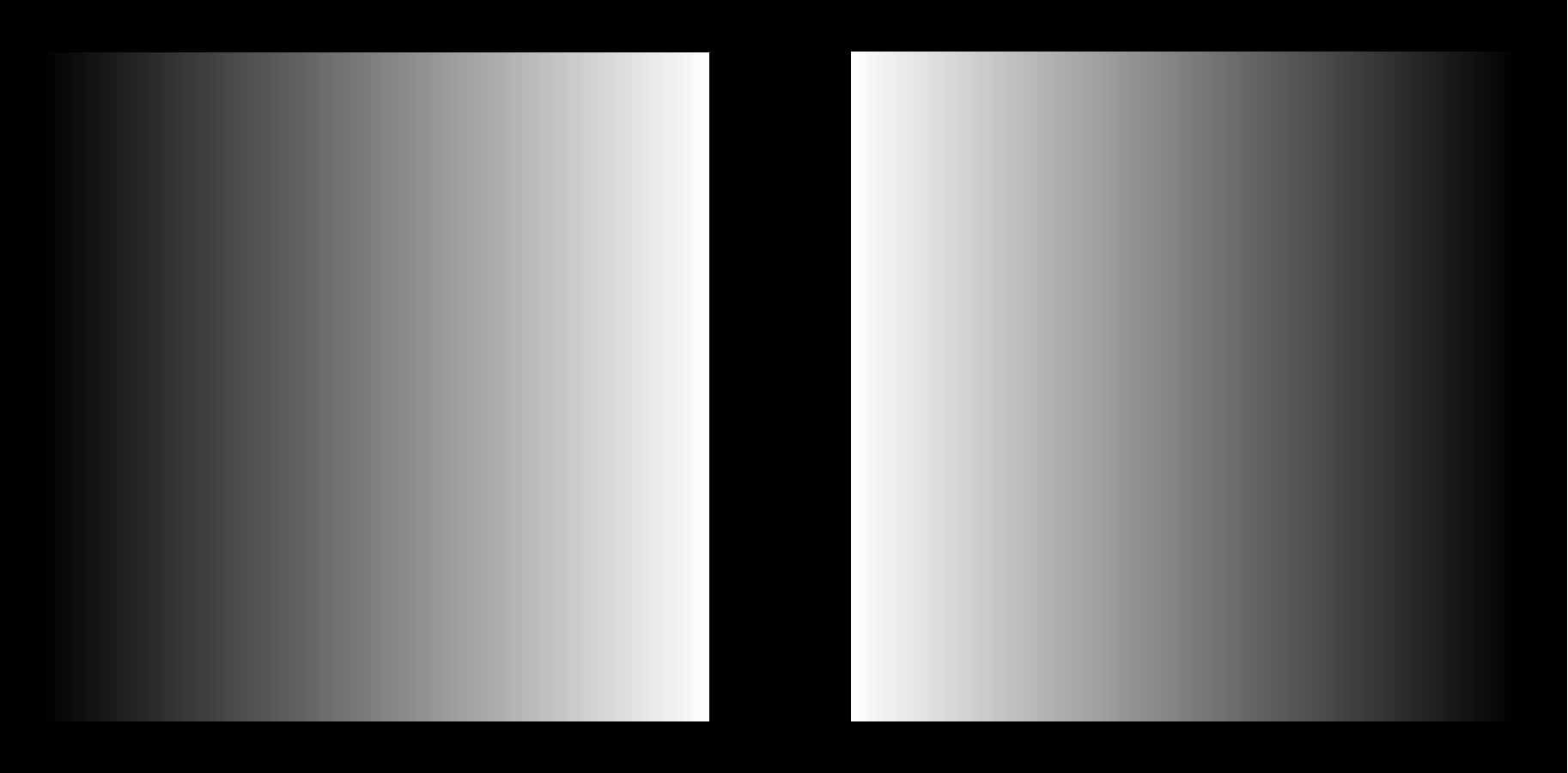}
         \caption{Ramp PSF}
     \end{subfigure}

    \caption{A comparison between different PSFs used to render the NIMAT effect. The two PSFs (i.e., DP PSF and Ramp PSF) are able to render smooth image motion. However, different motion transitions and artifacts can be introduced by using different PSFs. \textbf{Note: the synthetic output views are animated; click on the image to start the animation. It is recommended to open this PDF in Adobe Acrobat Reader to work properly.}
    }
    \label{fig:diff_psfs}
\end{figure}

\subsection{Results Using Other PSFs\label{subsec:other-psfs}}
As mentioned earlier in Sec.~\ref{sec:dpPSF}, the directional PSF used to render the NIMAT effect can be any generic PSF that involves blurring and shifting the image content. In Fig.~\ref{fig:diff_psfs}, we show the NIMAT effect rendered using two different PSF shapes -- namely, DP-based PSF (Fig.~\ref{fig:diff_psfs}, c) and transitional blurring 2D ramp mask with a constant intensity fall-off towards the opposite direction (i.e., Ramp PSF in Fig.~\ref{fig:diff_psfs}, d). These results demonstrate that other non-DP-based PSF can be utilized to render the NIMAT effect as long as it satisfies the conditions of having a transnational and blurring operator. Nevertheless, different motion transitions and artifacts can be introduced by using different PSFs as shown in Fig.~\ref{fig:diff_psfs}.
\section{Conclusion}
In this work, we proposed a modification to the DoF synthesis associated with most smartphones' {\it portrait mode} feature. This modification can be easily integrated into the traditional DoF synthesis unit and enables the generation of multiple sub-aperture views along with the synthetic bokeh photo. With this modification, we are also able to produce an aesthetic image motion effect similar to the novel NIMAT effect from~\cite{abuolaim2021multi}. For our multi-view synthesis, we introduced the novel idea of convolving the input image with the rotated blurring kernels based on the DoF blur and DP image formation. We validated our approach qualitatively and demonstrated that it produces smooth motion transition in the NIMAT effect with fewer artifacts compared to others. We aim to encourage work in this new research direction that presented a new pleasing effect of image motion.

\small
\bibliographystyle{ieee_fullname}
\bibliography{synth_dp_effect}
\end{document}